\documentclass[conference]{IEEEtran}
\IEEEoverridecommandlockouts
\usepackage{cite}
\usepackage{amsmath,amssymb,amsfonts}
\usepackage{graphicx}
\usepackage{textcomp}
\usepackage{xcolor}
\usepackage{algorithm}  
\usepackage{algpseudocode}

\usepackage{color}
\usepackage{float}
\usepackage{multirow}
\usepackage{booktabs}
\usepackage{multirow}
\usepackage{listings}
\usepackage{setspace}
\usepackage{makecell}
\usepackage{threeparttable}
\usepackage[inkscapelatex=false]{svg}
\usepackage{threeparttable}    
\def\BibTeX{{\rm B\kern-.05em{\sc i\kern-.025em b}\kern-.08em
    T\kern-.1667em\lower.7ex\hbox{E}\kern-.125emX}}
\begin{document}

\title{Research on Audio-Visual Quality Assessment Dataset and Method for User-Generated Omnidirectional Video}



\author{\IEEEauthorblockN{Fei Zhao, Da Pan*\thanks{* Corresponding Author}, Zelu Qi, Ping Shi}
\IEEEauthorblockA{\textit{School of Information and Communication Engineering, Communication University of China} \\
pigmo\_cuc@mails.cuc.edu.cn, \{pdmeng, theoneqi2001, shiping\}@cuc.edu.cn}}

\maketitle
\UseRawInputEncoding

\begin{abstract}
In response to the rising prominence of the Metaverse, omnidirectional videos (ODVs) have garnered notable interest, gradually shifting from professional-generated content (PGC) to user-generated content (UGC). However, the study of audio-visual quality assessment (AVQA) within ODVs remains limited. To address this, we construct a dataset of UGC omnidirectional audio and video (A/V) content. The videos are captured by five individuals using two different types of omnidirectional cameras, shooting 300 videos covering 10 different scene types. A subjective AVQA experiment is conducted on the dataset to obtain the Mean Opinion Scores (MOSs) of the A/V sequences. After that, to facilitate the development of UGC-ODV AVQA fields, we construct an effective AVQA baseline model on the proposed dataset, of which the baseline model consists of video feature extraction module, audio feature extraction and audio-visual fusion module. The experimental results demonstrate that our model achieves optimal performance on the proposed dataset.

\end{abstract}

\begin{IEEEkeywords}
User-Generated Omnidirectional Video, Audio-Visual Quality Assessment, Dataset Construction, Baseline Model
\end{IEEEkeywords}

\section{INTRODUCTION}

\begin{figure*}[!t]
	\centering
	\includegraphics[width=0.9 \textwidth]{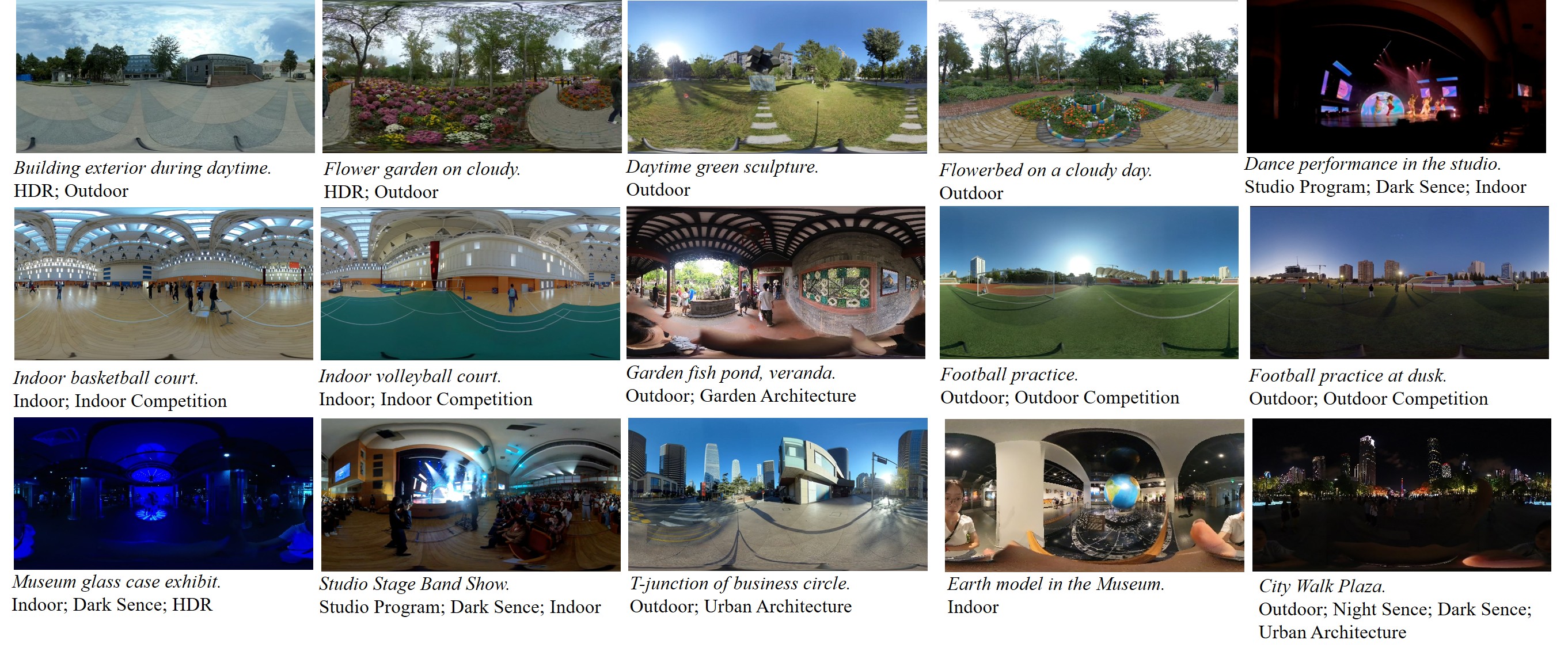}
	\caption{\textbf{Typical Scene Display of ODV Materials in the Dataset.} This figure provides examples of typical scenes from the ODV materials collected in our dataset, showing thumbnails of the typical video materials in ERP format. Additionally, it outlines the content of each scene and the corresponding scene category recorded in the videos.}\label{fig2}
\end{figure*}

As virtual reality applications proliferate and the underlying technologies evolve, omnidirectional videos (ODVs) have emerged as a vital medium, garnering significant attention from both industry and academic circles\cite{xu2020state}. In recent years, platforms for shooting, editing, and sharing ODVs aimed at average users have improved significantly, and ODVs have gradually become a part of everyday life\cite{wen2024perceptual}. The categories of ODV content are progressively transitioning from professional-generated content (PGC) to user-generated content (UGC). To optimize the viewing experience of ODVs for users\cite{bosman2024effect}, it is essential to conduct in-depth research on the quality assessment of UGC-ODV.

The construction of ODV datasets and the collection of subjective data are essential foundations for understanding how humans perceive the quality of ODVs and for improving objective assessment methods. Many research teams have developed datasets to assess the quality of ODVs, focusing on various tasks, such as VQA-ODV\cite{li2018bridge}, SSV360\cite{elwardy2022ssv360}, VRVQW\cite{wen2024perceptual}, and D-SAV350\cite{bernal2023d}. For various ODV application scenarios, each team has developed objective quality assessment methods tailored to their specific tasks, including OVQA-CNN\cite{li2018bridge}, OAVQA\cite{zhu2023perceptual}, ProVQA\cite{yang2022blind}, CIQNet\cite{hu2024omnidirectional}. These methods have shown promising results, encouraging further exploration of ODV quality assessment. However, most previous methods are based on PGC-ODVs and primarily focus on the video content, often neglecting the audio elements.

In summary, current subjective and objective quality assessment methods mainly focus on PGC-ODVs, emphasizing video quality while neglecting research on how audio quality affects human perception. ODVs offer users an immersive viewing experience through visual and auditory stimulation. This paper presents a UGC-ODV audio-visual quality dataset and proposes a baseline model to assess the audio-visual quality of ODV based on the dataset. The main contributions of this paper are as follows:
\begin{itemize}
    \item \textbf{A dataset for the audio-visual quality assessment of UGC-ODVs. }We construct a dedicated dataset for the assessment of the audio-visual quality of UGC-ODVs, including 300 ODV sequences that incorporate a range of audio-visual scenes and natural distortions, as well as their Mean Opinion Scores (MOSs). These sequences were captured using mainstream omnidirectional cameras, including Insta360 Pro 2 and Insta360 X3. We conducted a detailed and comprehensive analysis of the subjective dataset, resulting in 5,026 credible subjective quality assessment scores, alongside more than 12,06 million precise data entries related to head movements. To the best of our knowledge, this is the first work focusing on audio-visual quality assessment for UGC-ODV.
    \item \textbf{A baseline model for the UGC-ODV audio-visual quality assessment.} Utilizing the proposed dataset, we introduce an innovative no-reference method for audio-visual quality assessment. This approach incorporates modules for extracting visual and audio quality features, alongside a module for the fusion of multi-modal quality features, aimed at accurately predicting the audio-visual quality of UGC-ODVs. The experimental results validate the efficacy of the proposed model and offer insight for future research advancements.
\end{itemize}

\section{PROPOSED DATASET}
\subsection{UGC-ODV Audio-Visual Material Collection}
\begin{figure}[h]
	\centering
	\includegraphics[width=0.95\columnwidth]{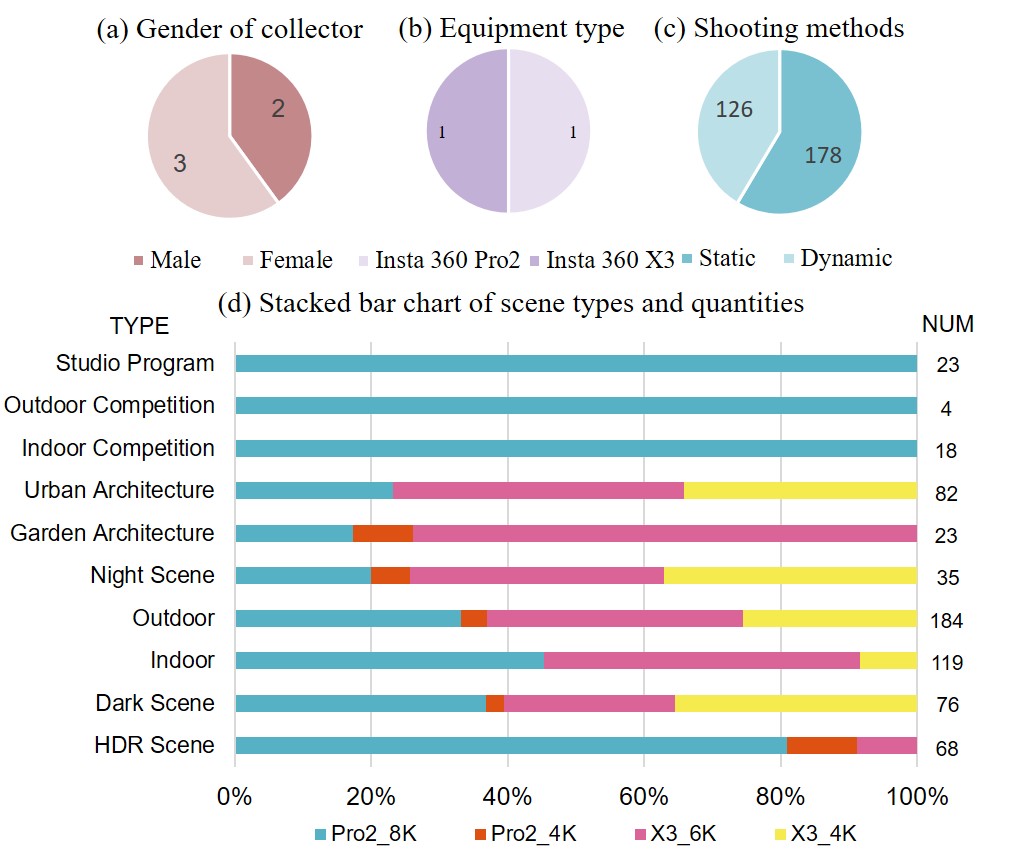}
	\caption{\textbf{Basic Statistics of Dataset Attributes.} This dataset was collected by two men and three women using two omnidirectional shooting devices for both static and dynamic recordings. It includes resolutions of 8K, 6K, and 4K and comprises 300 audio and video materials across ten types of scenes. The 10 types of scenes are: HDR Scene, Dark Scene, Indoor, Outdoor, Night Scene, Garden Architecture, Urban Architecture, Indoor Competition, Outdoor Competition, and Studio Program.}\label{fig1}
\end{figure}

\begin{figure}[htbp]
	\centering
	\includegraphics[width=0.8\columnwidth]{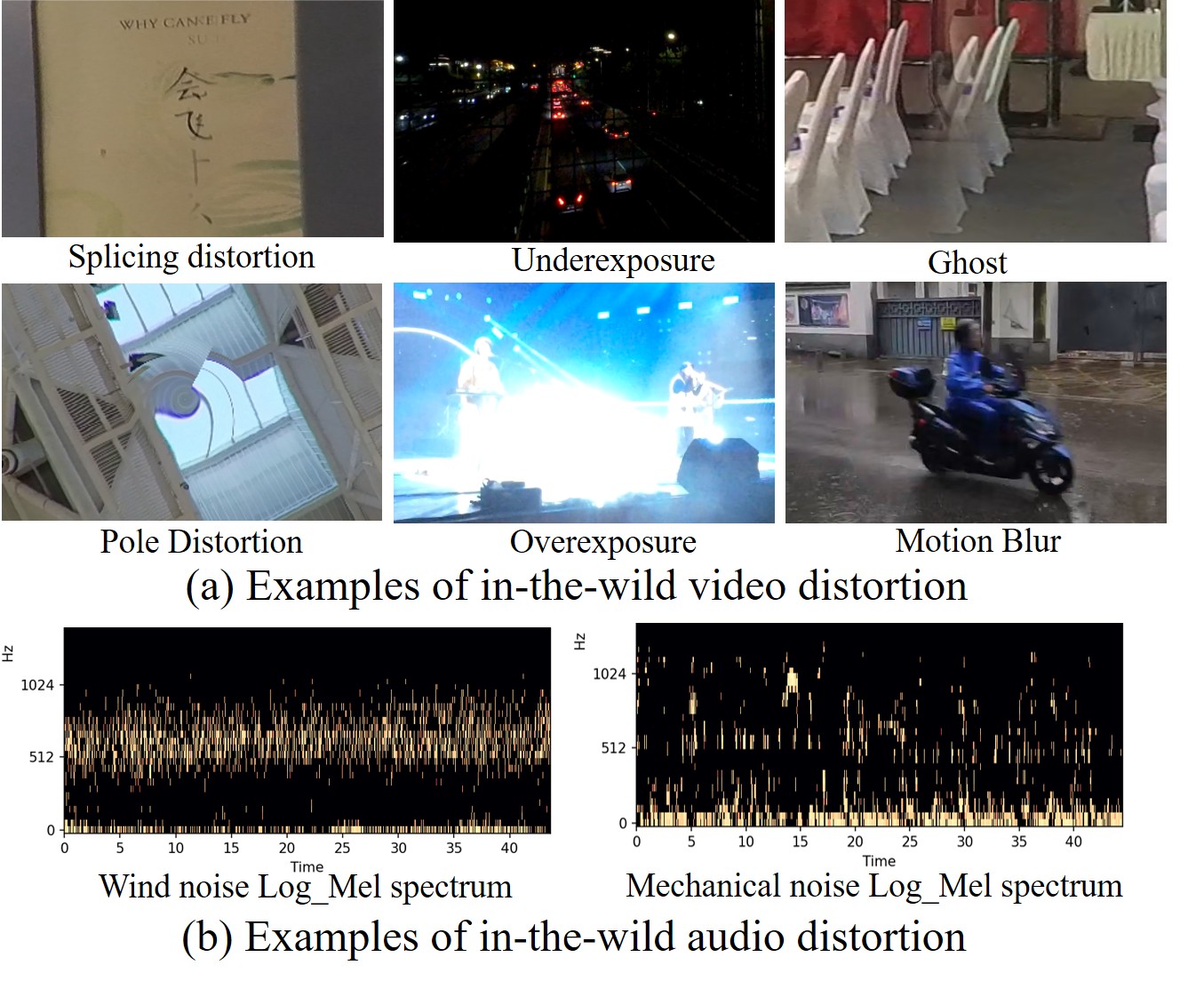}
	\caption{\textbf{Examples of Typical In-the-Wild Video and Audio Distortion Types in This Dataset.} This figure shows the typical types of video and audio distortion in the dataset constructed in this paper. These distortions are commonly encountered by users during everyday shooting.}\label{fig3}
\end{figure}

We simulate the shooting concepts of average users to capture and organize audio and video (A/V) materials, ultimately building a dataset of 300 UGC omnidirectional A/V materials encompassing 10 types of scenes. Basic statistical summaries and representative scene examples are depicted in Fig.~\ref{fig2} and Fig.~\ref{fig1}. Concurrently, our dataset includes prevalent A/V distortions observed in UGC omnidirectional content, such as ghosting, overexposure, motion blur, and wind noise, mechanical noise, as detailed in Fig.~\ref{fig3}.

\begin{figure}[htbp]
	\centering
	\includegraphics[width=0.8\columnwidth]{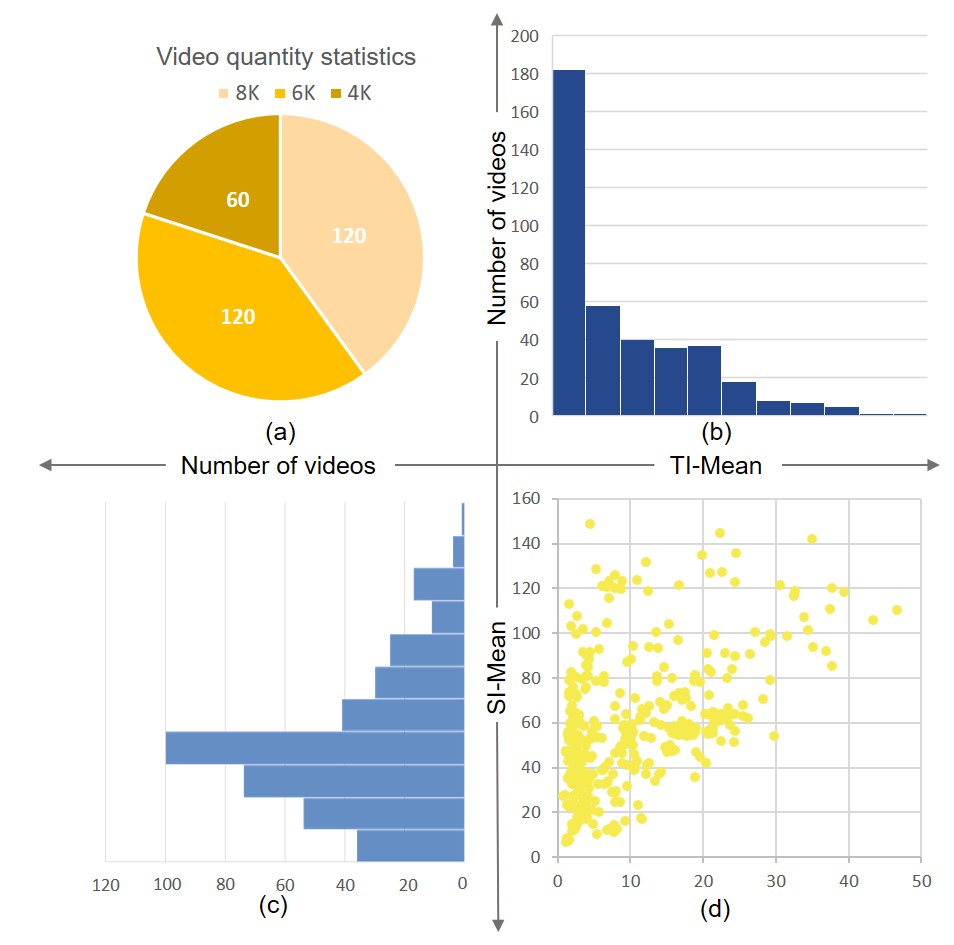}
	\caption{\textbf{Statistical Analysis of Spatial Information (SI) and Temporal Information (TI) in Video Materials.} \textbf{(a)} Pie chart of the total number of videos and resolution: shows that the number of video materials at each resolution is sufficient; \textbf{(b)} TI mean histogram: the TI mean is relatively low because the changes in the video image are restricted during the acquisition process to avoid causing discomfort to the audience and to ensure the accuracy of the quality ratings; \textbf{(c)} SI mean histogram: the SI mean is evenly distributed, indicating that the video materials are rich in spatial information and have diverse scenes; \textbf{(d)} Scatter plot of TI mean and SI mean: the wide distribution further confirms the diversity of the video materials.
}\label{fig4}
\end{figure}

To assess the diversity of the dataset, we perform a statistical assessment of the Spatial Information (SI) and Temporal Information (TI) metrics of the A/V data, illustrated in Fig.~\ref{fig4}. The analysis reveals that the SI and TI values span an extensive range, which corroborates the diverse nature of the A/V content within the dataset.

\subsection{Subjective Testing}
\begin{figure}[htbp]
	\centering
	\includegraphics[width=0.75\columnwidth]{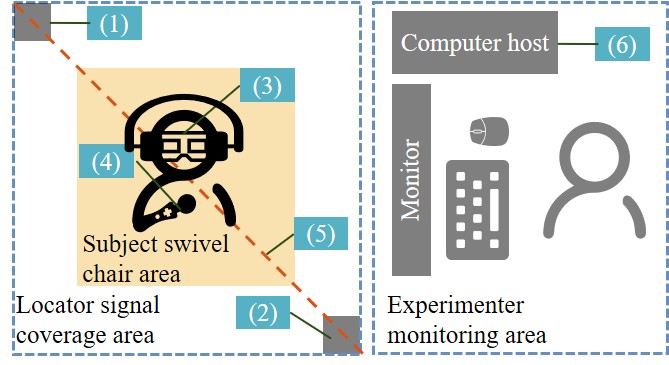}
	\caption{\textbf{Layout of the Subjective Experiment Space.} The layout positions two locators diagonally, with the subject's seat centered to maximize locator coverage and improve positioning accuracy. An experimenter's monitoring area ensures proper guidance and smooth progression of the experiment. The core hardware of the test environment includes: (1) HTC VIVE PRO2 locator; (2) HTC VIVE PRO2 locator; (3) HTC VIVE PRO2 head-mounted display; (4) HTC VIVE PRO2 control handle; (5) The diagonal of the space where the locators are positioned; (6) Host: GPU: NVIDIA GeForce RTX 4080; CPU: 13th Gen Intel(R) Core(TM) i5-13600KF.}\label{fig5}
\end{figure}

Since there are no established international standards for the subjective assessment of omnidirectional A/V content, we follow ITU-T P.910\cite{ITU-T.P.910} for the subjective assessment of ODVs and consider the application requirements of the HTC Vive Pro 2 head-mounted display to design the test environment and select participants for the subjective experiment. For the setup of the test environment, this project arranged a complete set of test hardware in a quiet indoor space, as shown in Fig.~\ref{fig5}, to ensure that the subjective experiment was conducted smoothly. We further design a user interface whereby the subjects could watch/listen and rate the ODVs. A continuous quality rating bar on the user interface was presented to the subject. After each A/V sequence was viewed, a continuous quality rating bar on the user interface was presented to the subject, which would be labeled with five Likert adjectives: Bad, Poor, Fair, Good, and Excellent. All subjects were instructed to give an opinion score on the overall A/V quality they perceived.

\begin{figure*}[h]
	\centering
	\includegraphics[width=0.85\textwidth]{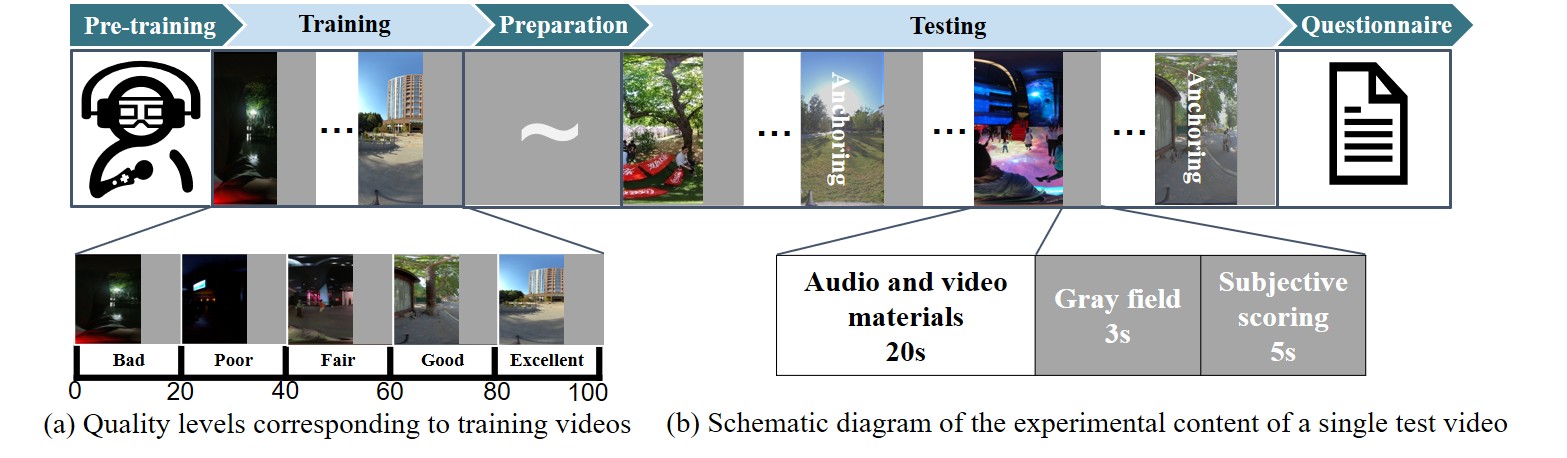}
	\caption{\textbf{Schematic Diagram of Subjective Experiment Process.} \textbf{Pre-training:} The subjects sit in a swivel chair and get used to the viewing environment and equipment. They then wear the display device and learn to use the handle. If they feel uncomfortable, the test is terminated. \textbf{Training:} The subjects read the experimental purpose and scoring rules before taking the test and then proceed with the test. If they feel uncomfortable, the test is terminated. \textbf{Preparation:} Subjects prepare for formal experiment. \textbf{Testing:} The subjects will watch and score 33 20 seconds UGC-ODV audio-visual materials, including 30 test videos, and 3 anchor videos, while providing head movement data and subjective scores. \textbf{Questionnaire:} The subjects will fill out the Simulator Discomfort Questionnaire (SSQ)\cite{balk2013simulator} and the Group Presence Questionnaire (IPQ)\cite{IPQ} to quantify the feeling of dizziness and immersion in the experiment and prepare for subsequent data processing. \textbf{Anchoring:} A ``Good" quality anchor video is played after every 10 test videos to check scoring consistency and reduce fatigue. In \textbf{(a)}, five training videos are screened with reference to the continuous rating scale, with the following conditions: \textbf{Bad:} 4K resolution, severe shaking, poor stability, unclear details, and noisy sound. \textbf{Poor:} 8K resolution, stable fixed lens, extremely poor clarity, unclear details, and noisy sound. \textbf{Fair:} 6K resolution, slightly shaking moving lens, visible in most details, and relatively clear sound. \textbf{Good:} 8K resolution, stable fixed lens, overexposed just light area, and clear sound. \textbf{Excellent:} 8K resolution, stable fixed lens, completely clear details, and clear sound.  In \textbf{(b)}, each test includes 20 seconds of video, 3 seconds of a gray field, and 5 seconds for subjective rating, with the gray field helping to relieve the subjects' visual fatigue.}\label{fig6}
\end{figure*}

The subjective experiment gathers data comprising opinion scores and head movement information. The coordinates of the head movement are obtained at a sampling rate of 120Hz, capturing three angles: pitch, yaw, and rotation. Following the guidelines of the ITU-T BT.500 standard\cite{ITU-R.BT.500-13}, a minimum of 18 individuals with normal sight and hearing participate in the viewing and scoring of each video. Figure~\ref{fig6} illustrates the design of the experiment. We employ the Single Stimulus Continuous Quality Evaluation (SSCQE)\cite{ITU-R.BT.500-13} technique to collect both quality ratings and head movement metrics from participants during the audio-visual assessment.
In terms of subjective experimental process design, this project divides the 300 test sequences into 10 groups and selects 5 training videos and 3 anchor videos from the redundant videos based on four factors: video resolution, video stability, overall clarity and sound source clarity, as shown in Fig.~\ref{fig6}. The entire subjective experiment is primarily divided into five stages. See Fig.~\ref{fig6} for details.

A total of 136 subjects participated in the subjective experiment, completing 190 assessment groups and providing 5,700 subjective scores. Among them, 36 subjects participated in multiple sessions with at least two days between experiments and different test content each time. The subjects were mainly between 18 and 27 years old, with 43 males and 95 females. All subjects met the following criteria: normal or corrected-to-normal vision and color vision; normal hearing; the ability to watch ODVs without discomfort, and no prior experience in ODV quality assessment.

\subsection{Subjective Data Processing and Summary}

We follow the ITU-T BT.500\cite{ITU-R.BT.500-13} standard to process and analyze quality rating data obtained from subjective experiments, ensuring a reliable and accurate Mean Opinion Score (MOS). Since dizziness may influence subjective scores, this paper employs the Simulator Sickness Questionnaire (SSQ) to quantify dizziness levels. Subjective scores associated with abnormal SSQ values are identified as outliers and excluded to ensure data accuracy.

\begin{figure}[H]
	\centering
	\includegraphics[width=0.5\columnwidth]{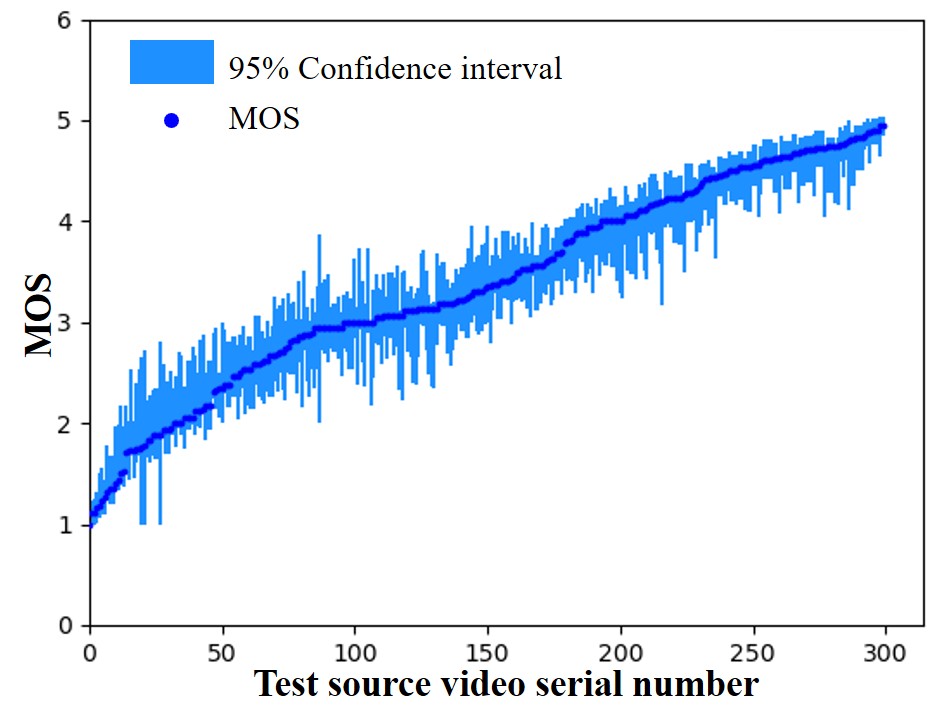}
	\caption{\textbf{MOSs and 95\% Confidence Intervals for Each Video.}}\label{fig9}
\end{figure}

According to the data processing recommendations given in ITU-T BT.500\cite{ITU-R.BT.500-13}, the remaining subjective scoring data of subjects were screened for valid scores with a confidence interval 95\%, and MOS was calculated. After screening, the number of valid ratings remaining within the 95\% confidence interval for each test video is greater than or equal to 15, meeting the requirements of ITU-T BT.500\cite{ITU-R.BT.500-13}. Based on the valid scores obtained, the MOSs of the test videos are calculated. As shown in Fig.~\ref{fig9}, the results showed that all MOS fell within the corresponding 95\% confidence interval, verifying the effectiveness of MOSs.

In the collection and processing of head movement data, the validity of the subject's subjective rating reflects the degree of their seriousness in watching. Therefore, this paper believes that the head movement data corresponding to the valid rating is also valid and does not require additional screening.

We compare our dataset with four mainstream datasets in ODV research, as shown in Table~\ref{tbl2}. Overall, this dataset offers a rich array of data sources, ample data volume, rigorous and reliable subjective experiments, and holds significant theoretical research value.

\renewcommand{\thetable}{1}
\begin{table}[h]
\normalsize
\centering
\caption{dataset indicator comparison table.}\label{tbl2}
\begin{threeparttable}          
\scalebox{0.8}{
\begin{tabular}{c|c|c|c}
\toprule
\textbf{Dataset}              & \textbf{Num}\tnote{1}  & \textbf{Resolution}                                           & \textbf{HM/EM/MOS} \\ \midrule
VQA-ODV\cite{li2018bridge}    & 60/540                 & \begin{tabular}[c]{@{}c@{}}3840×1920\\ 7680×3840\end{tabular} & HM+EM+MOS          \\
VOD-VQA\cite{xu2020viewport}  & 4/120                  & 2K$\sim$8K                                                    & EM                 \\
VRVQW\cite{wen2024perceptual} & -/502                  & 1.5K$\sim$5.3K                                                & EM+MOS             \\
D-SAV360\cite{bernal2023d}    & -/85                   & 3840×1920                                                     & HM+EM              \\
\textbf{Ours}                 & \textbf{-/300}         & \textbf{4K$\sim$8K}                                           & \textbf{HM+MOS}    \\ \midrule
\textbf{Dataset}              & \textbf{\begin{tabular}[c]{@{}c@{}}Audio\\ Distortion\end{tabular}} & \textbf{\begin{tabular}[c]{@{}c@{}}Video\\ Distortion\end{tabular}} & \textbf{Source} \\ \midrule
VQA-ODV\cite{li2018bridge}    & N/A                                                                 & Synthetic                                                           & Internet                                                                      \\
VOD-VQA\cite{xu2020viewport}  & N/A                                                                 & Synthetic                                                           & Internet                                                                       \\
VRVQW\cite{wen2024perceptual} & N/A                                                                 & TRUE                                                                & Insta 360 X3                                                                   \\
D-SAV360\cite{bernal2023d}    & TRUE                                                                & TRUE                                                                & \begin{tabular}[c]{@{}c@{}}50 team shots\\ 35 from the internet\end{tabular}   \\
\textbf{Ours}                 & \textbf{TRUE}                                                       & \textbf{TRUE}                                                       & \textbf{\begin{tabular}[c]{@{}c@{}}Insta 360 X3\\ Insta 360 Pro2\end{tabular}} \\ \bottomrule
\end{tabular}
}
\begin{tablenotes}    
\footnotesize              
\item[1] Number of reference videos/ Number of distorted videos.
\end{tablenotes}            
\end{threeparttable}       
\end{table}

\begin{figure*}[h]
	\centering
	\includegraphics[width=0.82\textwidth]{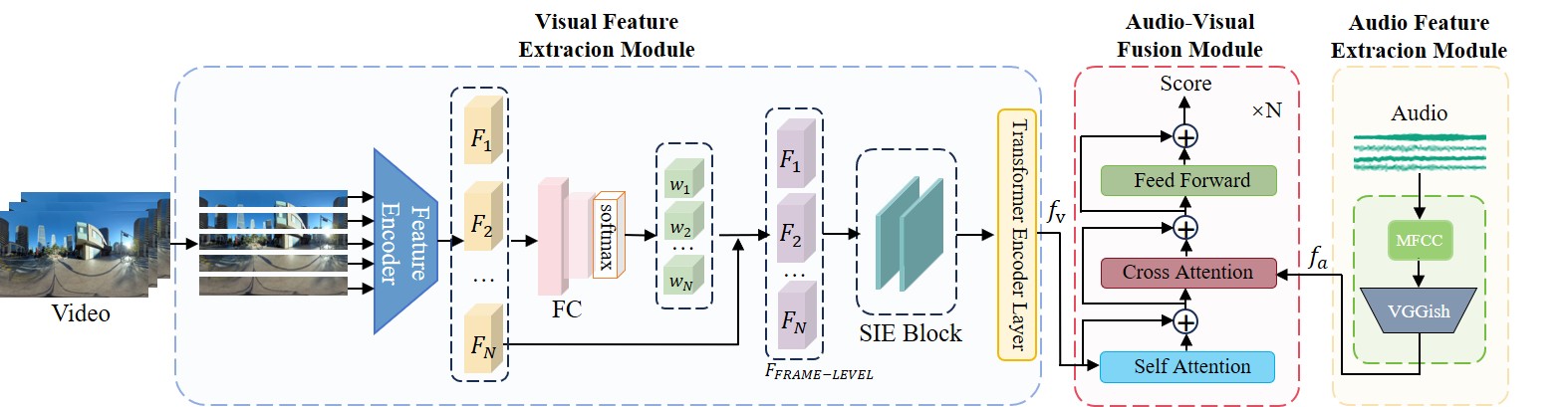}
	\caption{\textbf{Model Architecture.} It includes visual feature extraction module, audio feature extraction module and audio-visual feature fusion module.}\label{fig11}
\end{figure*}

\renewcommand{\thetable}{2}
\begin{table*}[h]
\centering
\caption{Model performance comparison table.}\label{tbl3}
\begin{threeparttable}          
\scalebox{0.8}{
\begin{tabular}{c|cccc|cccc}
\toprule
\multirow{2}{*}{\textbf{Approaches}} & \multicolumn{4}{c|}{\textbf{Attributes}}   & \multicolumn{4}{c}{\textbf{Performance on Our Dataset}} \\ \cmidrule{2-9} 
                                   &No reference& Finetuned & Type\textsuperscript{1}    & With Audio & SROCC$\uparrow$     & PLCC$\uparrow$     & KROCC$\uparrow$     & RMSE$\downarrow$    \\ \midrule
DOVER\cite{wu2023exploring}        & \checkmark & \checkmark & 2D video  & ×          & 0.7974    & 0.7929   & 0.5993   & 0.6024   \\
FastVQA\cite{wu2022fast}           & \checkmark & \checkmark & 2D video  & ×          & 0.7256    & 0.7280   & 0.5232   & 0.6903   \\
CIQNet\cite{hu2024omnidirectional} & \checkmark & \checkmark & ODV       & ×          & 0.8045    & 0.8254   & 0.6186   & 0.7198   \\
ProVQA\cite{yang2022blind}         & \checkmark & \checkmark & ODV       & ×          & 0.8081    & 0.8204   & 0.6390   & 0.5574   \\ \midrule
Ours (Without Audio)\tnote{*}      & \checkmark &\checkmark  & ODV       & ×          & 0.8045    & 0.8254   & 0.6186   & 0.7198   \\
Ours (Cat)\tnote{*}                & \checkmark &\checkmark  & ODV       &\checkmark  & 0.7996    & 0.8262   & 0.6129   & 0.7366   \\
Ours (Add)\tnote{*}                & \checkmark &\checkmark  & ODV       &\checkmark  & 0.7921    & 0.8234   & 0.5981   & \textbf{0.5339}   \\ \midrule
\textbf{Ours}                      & \checkmark &\checkmark  & ODV       &\checkmark  & \textbf{0.8245}    & \textbf{0.8590}   & \textbf{0.6436}   & 0.5772   \\  \bottomrule
\end{tabular}
}
\begin{tablenotes}    
\footnotesize              
\item[*] Represent the results of ablation experiments, including the ablation of the presence or absence of audio branches and feature fusion methods.
\item[1] Represent whether the method is applied to evaluate 2D planar videos or ODVs.
\end{tablenotes}            
\end{threeparttable}       
\end{table*}

\section{PROPOSED ODV-AVQA BASELINE MODEL}

Based on the dataset we constructed, we further propose a no-reference UGC-ODV audio-visual objective quality assessment model. The model includes three basic modules, namely the visual feature extraction module, the audio feature extraction module, and the audio-visual fusion module, as shown in Figure 8. The details of each module will be described in detail in the following.

\subsection{Video Feature Extraction Module}

Hu et al.\cite{hu2024omnidirectional} introduced a Causal Intervention-based Quality
prediction Network (CIQNet) to address the causal impact of dimensional confounding factors on video quality feature extraction in ODVs, achieving outstanding performance. We adopt the basic structure of CIQNet to extract visual features. The network first partitions the ODV into sub-regions based on dimensions and trains feature encoders for each sub-region to achieve dimension-invariant representations, thereby mitigating the influence of confounding factors on feature representation. Subsequently, a backdoor adjustment module is employed to further eliminate the relationship between dimensional confounding factors and video content. This is accomplished by estimating latitude-related weights for each sub-region and using these weights to aggregate features extracted from different sub-regions. Finally, the temporal dependencies of the aggregated features are modeled to produce comprehensive video quality features $f_v$.

\subsection{Audio Feature Extraction Module}

Given that VGGish has good audio feature extraction capabilities and has been widely used in the field of audio feature extraction tasks\cite{diwakar2023robust},\cite{geng2023dense},\cite{li2023catr}, this paper uses the VGGish network to extract features from the audio of ODVs. This module inputs the original audio into the MFCC module to obtain a stable Mel spectrogram, and then inputs the stable Mel spectrogram into the VGGish model consisting of four convolutional layers and one pooling layer to finally obtain the audio feature $f_{a}$.

\subsection{Audio-Visual Fusion Module}

After having the features from perspectives of video quality and audio fidelity, we design a transformer-based fusion module to fuse those two features. The module includes $N$ blocks with self-attention, cross-attention, and feed-forward layers in each block. Initially, video features $f_{v}$ are processed through self-attention layers, followed by interaction with the audio features $f_{a}$ in cross-attention layers (as applied in every alternate transformer block). The fusion module enables the model to coherently integrate features from both domains, facilitating a more thorough comprehension of the audio-visual characteristics.

\section{EXPERIMENT}

\subsection{Experiment Protocol}

\subsubsection{Benchmark Dataset}
The 300 UGC-ODV datasets were randomly split into training and test sets at a ratio of 80\% to 20\%. The videos are stored in ERP projection format, each with a duration of 20 seconds. Resolutions cover 4K, 6K, and 8K, and the audio includes two-channel and four-channel formats.

\subsubsection{Assessment Metrics}
We use four criteria to assess model performance: Pearson linear correlation coefficient (PLCC), Spearman rank correlation coefficient (SROCC), Kendall rank correlation coefficient (KROCC), and Root mean square error (RMSE). Among them, SROCC and KROCC measure the monotonicity of prediction, and PLCC and RMSE measure the accuracy of prediction. The larger the PLCC, SROCC, and KROCC values, the higher the correlation; the smaller the RMSE value, the lower the prediction error. Before calculating these indicators, the predicted scores are fitted with the corresponding true values using a logistic function so that the proportion of the fitted scores is consistent with the true values.

\subsubsection{Compared Methods}
We compare our proposed method with the following models: DOVER\cite{wu2023exploring}, Fast-VQA\cite{wu2022fast}, CIQNet\cite{hu2024omnidirectional}, ProVQA\cite{yang2022blind}. At the same time, we remove the audio branch of the baseline model and replace the Transformer blocks of the fusion module with Add and Cat for ablation comparison.

\subsubsection{Experiment Setup}

Our method is implemented based on the PyTorch framework and the Tensorflow framework and runs on four NVIDIA GeForce RTX 3090 GPUs with 24GB of memory. The pre-training parameters of the ODV feature extraction network are based on the pre-training data obtained on the IQA-ODI\cite{Li2021Spatial}. The pre-training parameters of the audio feature extraction network are based on YouTube-8M\cite{abu2016youtube}. During the training phase, the baseline model uses the Adam optimizer with a learning rate set to 1e-3. Other models also use the Adam optimizer during fine-tuning with a learning rate set to 3e-5.

\subsection{Performance Comparison}

\begin{figure}[h]
	\centering
	\includegraphics[width=0.96 \columnwidth]{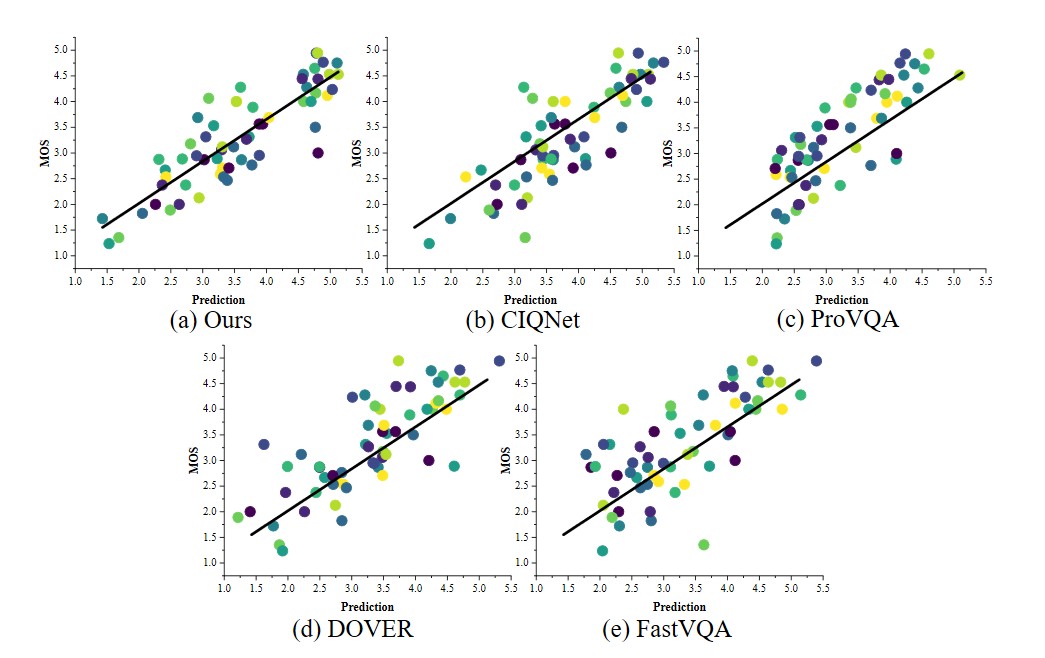}
	\caption{\textbf{Scatter Plots of the MOS and Prediction.} A four-order polynomial nonlinear fitting obtains the curves. The brightness of scatter points from dark to bright means density from low to high.}\label{fig12}
\end{figure}

We select two of the most advanced ODV no-reference quality assessment methods and two of the most commonly used UGC-VQA methods for performance testing.
As can be seen from Table~\ref{tbl3} and Fig.~\ref{fig12}, our model achieves SOTA performance, while the CIQNet\cite{hu2024omnidirectional} and ProVQA\cite{yang2022blind} show especially serious performance degradation. This is because UGC-ODVs have a variety of video scene types, video resolutions, shooting devices and the inability of users to take into account both video and audio quality when shooting. These factors bring huge challenges to the no-reference objective quality assessment, leading to pronounced performance drops in previously optimal methods. Other UGC-VQA methods perform poorly because they didn't consider the special spherical properties of ODVs. Therefore, the task of no-reference Audio-Visual quality assessment under UGC-ODVs still needs further exploration and in-depth research.

In the ablation experiment phase, we conduct tests by removing the audio feature extraction module and replacing the Transformer with Add and Cat operations, as shown in Table~\ref{tbl3}. The experimental results show that both the lack of audio branch and the substitution of feature fusion methods can cause obvious performance degradation, which shows that 1) it is meaningful to consider the impact of audio on quality perception in the UGC-ODV quality assessment task; 2) selecting the correct fusion method is crucial to model performance. 

\section{CONCLUSION}
In this paper, to enhance the audio-visual experience of ordinary users in ODVs, we conducted a study on the quality assessment of user-generated ODV content. Addressing the limitations of existing datasets, we constructed a large-scale dataset of user-generated ODVs, collecting perceptual quality scores and head movement data through subjective experiments. Based on this dataset, we developed and implemented a no-reference audio-visual quality assessment model for ODVs, demonstrating its effectiveness. We will investigate deeper the mechanism of audio-visual collaborative perception and explore more efficient quality assessment methods to improve model performance. \par

%
\bibliographystyle{IEEEtran}
\bibliography{IEEEabrv,ref}

\begin{thebibliography}{10}
\providecommand{\url}[1]{#1}
\csname url@samestyle\endcsname
\providecommand{\newblock}{\relax}
\providecommand{\bibinfo}[2]{#2}
\providecommand{\BIBentrySTDinterwordspacing}{\spaceskip=0pt\relax}
\providecommand{\BIBentryALTinterwordstretchfactor}{4}
\providecommand{\BIBentryALTinterwordspacing}{\spaceskip=\fontdimen2\font plus
\BIBentryALTinterwordstretchfactor\fontdimen3\font minus \fontdimen4\font\relax}
\providecommand{\BIBforeignlanguage}[2]{{%
\expandafter\ifx\csname l@#1\endcsname\relax
\typeout{** WARNING: IEEEtran.bst: No hyphenation pattern has been}%
\typeout{** loaded for the language `#1'. Using the pattern for}%
\typeout{** the default language instead.}%
\else
\language=\csname l@#1\endcsname
\fi
#2}}
\providecommand{\BIBdecl}{\relax}
\BIBdecl

\bibitem{xu2020state}
M.~Xu, C.~Li, S.~Zhang, and P.~Le~Callet, ``State-of-the-art in 360 video/image processing: Perception, assessment and compression,'' \emph{IEEE Journal of Selected Topics in Signal Processing}, vol.~14, no.~1, pp. 5--26, 2020.

\bibitem{wen2024perceptual}
W.~Wen, M.~Li, Y.~Yao, X.~Sui, Y.~Zhang, L.~Lan, Y.~Fang, and K.~Ma, ``Perceptual quality assessment of virtual reality videos in the wild,'' \emph{IEEE Transactions on Circuits and Systems for Video Technology}, 2024.

\bibitem{bosman2024effect}
I.~d.~V. Bosman, O.~â. Buruk, K.~J{\o}rgensen, and J.~Hamari, ``The effect of audio on the experience in virtual reality: a scoping review,'' \emph{Behaviour \& Information Technology}, vol.~43, no.~1, pp. 165--199, 2024.

\bibitem{li2018bridge}
C.~Li, M.~Xu, X.~Du, and Z.~Wang, ``Bridge the gap between vqa and human behavior on omnidirectional video: A large-scale dataset and a deep learning model,'' in \emph{Proceedings of the 26th ACM international conference on Multimedia}, 2018, pp. 932--940.

\bibitem{elwardy2022ssv360}
M.~Elwardy, H.-J. Zepernick, and Y.~Hu, ``Ssv360: A dataset on subjetive quality assessment of 360 videos for standing and seated viewing on an hmd,'' in \emph{2022 IEEE Conference on Virtual Reality and 3D User Interfaces Abstracts and Workshops (VRW)}.\hskip 1em plus 0.5em minus 0.4em\relax IEEE, 2022, pp. 01--06.

\bibitem{bernal2023d}
E.~Bernal-Berdun, D.~Martin, S.~Malpica, P.~J. Perez, D.~Gutierrez, B.~Masia, and A.~Serrano, ``D-sav360: A dataset of gaze scanpaths on 360° ambisonic videos,'' \emph{IEEE Transactions on Visualization and Computer Graphics}, 2023.

\bibitem{zhu2023perceptual}
X.~Zhu, H.~Duan, Y.~Cao, Y.~Zhu, Y.~Zhu, J.~Liu, L.~Chen, X.~Min, and G.~Zhai, ``Perceptual quality assessment of omnidirectional audio-visual signals,'' in \emph{CAAI International Conference on Artificial Intelligence}.\hskip 1em plus 0.5em minus 0.4em\relax Springer, 2023, pp. 512--525.

\bibitem{yang2022blind}
L.~Yang, M.~Xu, S.~Li, Y.~Guo, and Z.~Wang, ``Blind vqa on 360° video via progressively learning from pixels, frames, and video,'' \emph{IEEE Transactions on Image Processing}, vol.~32, pp. 128--143, 2022.

\bibitem{hu2024omnidirectional}
Z.~Hu, L.~Liu, and Q.~Sang, ``Omnidirectional video quality assessment with causal intervention,'' \emph{IEEE Transactions on Broadcasting}, 2024.

\bibitem{ITU-T.P.910}
ITU, \emph{Subjective video quality assessment methods for multimedia applications}.\hskip 1em plus 0.5em minus 0.4em\relax ITU, 2008.

\bibitem{ITU-R.BT.500-13}
{ITU}, \emph{Methodology for the subjective assessment of the quality of television pictures}.\hskip 1em plus 0.5em minus 0.4em\relax ITU, 2012.

\bibitem{balk2013simulator}
S.~A. Balk, D.~B. Bertola, and V.~W. Inman, ``Simulator sickness questionnaire: twenty years later,'' in \emph{Driving Assessment Conference}, vol.~7, no. 2013.\hskip 1em plus 0.5em minus 0.4em\relax University of Iowa, 2013.

\bibitem{IPQ}
IPQ, ``Ipq,'' https://igroup.org/pq/ipq/index.php, 1 2016.

\bibitem{xu2020viewport}
M.~Xu, L.~Jiang, C.~Li, Z.~Wang, and X.~Tao, ``Viewport-based cnn: A multi-task approach for assessing 360° video quality,'' \emph{IEEE Transactions on Pattern Analysis and Machine Intelligence}, vol.~44, no.~4, pp. 2198--2215, 2020.

\bibitem{wu2023exploring}
H.~Wu, E.~Zhang, L.~Liao, C.~Chen, J.~Hou, A.~Wang, W.~Sun, Q.~Yan, and W.~Lin, ``Exploring video quality assessment on user generated contents from aesthetic and technical perspectives,'' in \emph{Proceedings of the IEEE/CVF International Conference on Computer Vision}, 2023, pp. 20\,144--20\,154.

\bibitem{wu2022fast}
H.~Wu, C.~Chen, J.~Hou, L.~Liao, A.~Wang, W.~Sun, Q.~Yan, and W.~Lin, ``Fast-vqa: Efficient end-to-end video quality assessment with fragment sampling,'' in \emph{European conference on computer vision}.\hskip 1em plus 0.5em minus 0.4em\relax Springer, 2022, pp. 538--554.

\bibitem{diwakar2023robust}
M.~Diwakar and B.~Gupta, ``The robust feature extraction of audio signal by using vggish model,'' 2023.

\bibitem{geng2023dense}
T.~Geng, T.~Wang, J.~Duan, R.~Cong, and F.~Zheng, ``Dense-localizing audio-visual events in untrimmed videos: A large-scale benchmark and baseline,'' in \emph{Proceedings of the IEEE/CVF Conference on Computer Vision and Pattern Recognition}, 2023, pp. 22\,942--22\,951.

\bibitem{li2023catr}
K.~Li, Z.~Yang, L.~Chen, Y.~Yang, and J.~Xiao, ``Catr: Combinatorial-dependence audio-queried transformer for audio-visual video segmentation,'' in \emph{Proceedings of the 31st ACM International Conference on Multimedia}, 2023, pp. 1485--1494.

\bibitem{Li2021Spatial}
L.~Yang, M.~Xu, X.~Deng, and B.~Feng, ``Spatial attention-based non-reference perceptual quality prediction network for omnidirectional images,'' in \emph{2021 IEEE International Conference on Multimedia and Expo (ICME)}, 2021, pp. 1--6.

\bibitem{abu2016youtube}
S.~Abu-El-Haija, N.~Kothari, J.~Lee, P.~Natsev, G.~Toderici, B.~Varadarajan, and S.~Vijayanarasimhan, ``Youtube-8m: A large-scale video classification benchmark,'' \emph{arXiv preprint arXiv:1609.08675}, 2016.

\end{thebibliography}
\end{document}